\documentclass[conference]{IEEEtran}

\IEEEoverridecommandlockouts
\usepackage{cite}
\usepackage{amsmath,amssymb,amsfonts}
\usepackage{algorithmic}
\usepackage{float}
\usepackage{graphicx}
\usepackage{enumitem}
\usepackage{caption}
\usepackage{subcaption}
\usepackage{textcomp}
\usepackage{xcolor}
\usepackage[]{hyperref}
\usepackage{booktabs,siunitx,tabularx, stfloats}
\usepackage{threeparttable}
\usepackage{svg}
\usepackage{diagbox}
\usepackage{fancyhdr}
\usepackage{bbding}
\usepackage{flushend}
\newcommand{\orcid}[1]{\href{https://orcid.org/#1}{\includesvg[width=10pt]{imgs/orcid}}}
\renewcommand{\baselinestretch}{0.9074}

\def\BibTeX{{\rm B\kern-.05em{\sc i\kern-.025em b}\kern-.08em
    T\kern-.1667em\lower.7ex\hbox{E}\kern-.125emX}}
\newcommand{\printfnsymbol}[1]{%
  \textsuperscript{\@fnsymbol{#1}}%
}

\usepackage{glossaries}
\newacronym{gan}{GAN}{Generative Adversarial Network}
\newacronym{ct}{CT}{Computed Tomography}
\newacronym{da}{DA}{Data Augmentation}
\newacronym{cnn}{CNN}{Convolutional Neural Network}
\newacronym{cv}{CV}{Computer Vision}
\newacronym{mbconv}{MBConv}{Mobile inverted Bottleneck Convolution}
\newacronym{nn}{NN}{Neural Network}
\newacronym{clahe}{CLAHE}{Contrast Limited Adaptive Histogram Equalization}
\newacronym{auc}{AUC}{Area Under the Curve}
\newacronym{ba}{BA}{Balanced Accuracy}
\newacronym{ddpm}{DDPM}{Denoising Diffusion Probabilistic Model}
\newacronym{dl}{DL}{Deep Learning}
\newacronym{ai}{AI}{Artificial Intelligence}

\newcommand{\I}{\mathbf{I}}
\newcommand{\Patch}{\mathbf{P}}

\begin{document}

\title{Localization of Synthetic Manipulations in \\Western Blot Images}

\author{Anmol Manjunath\textsuperscript{1}, Viola Negroni\textsuperscript{1}, Sara Mandelli\textsuperscript{1}, Daniel Moreira\textsuperscript{2}, Paolo Bestagini\textsuperscript{1}\\ 
\small{\textsuperscript{1}Dipartimento di Elettronica, Informazione e Bioingegneria, Politecnico di Milano, 20133 Milan, Italy.} \\
\small{\textsuperscript{2}Department of Computer Science, Loyola University Chicago, Chicago, IL, USA.} \\
Emails: viola.negroni@polimi.it, sara.mandelli@polimi.it\\
\thanks{This research is sponsored by the Defense Advanced
Research Projects Agency (DARPA) and the Air Force Research Laboratory
(AFRL) under agreement number FA8750-20-2-1004. 
This work
was partially supported by the European Union under the Italian National
Recovery and Resilience Plan (NRRP) of NextGenerationEU
(PE00000001 - program ``RESTART'' and PE00000014 - program ``SERICS'') and 
by the ``FOSTERER'' project, funded by the Italian Ministry of University and Research within PRIN 2022 program.}
}

\maketitle

 

\begin{abstract}
Recent breakthroughs in deep learning and generative systems have significantly fostered the creation of synthetic media, as well as the local alteration of real content via the insertion of highly realistic synthetic manipulations. 
Local image manipulation, in particular, poses serious challenges to the integrity of digital content and societal trust. 
This problem is not only confined to multimedia data, but also extends to biological images included in scientific publications, like images depicting Western blots. 
In this work, we address the task of localizing synthetic manipulations in Western blot images. 
To discriminate between pristine and synthetic pixels of an analyzed image, we propose a synthetic detector that operates on small patches extracted from the image. We aggregate patch contributions to estimate a tampering heatmap, highlighting synthetic pixels out of pristine ones.
Our methodology proves effective when tested over
two manipulated Western blot image datasets, one altered automatically and the other manually by exploiting advanced AI-based image manipulation tools that are unknown at our training stage. We also explore the robustness of our method over an external dataset of other scientific images depicting different semantics, manipulated through unseen generation techniques. 
We release our experimental code and the manipulated datasets at \href{https://github.com/polimi-ispl/western-blot-synthetic-manipulation-localization}{{https://github.com/polimi-ispl/western-blot-synthetic-manipulation-localization}}.
\end{abstract}


\section{Introduction}
\label{sec:intro}
In recent times, synthetic content generation methods have been gaining a lot of popularity. The rapid advancements of \gls{ai} over the past decade have incredibly simplified the creation of images exhibiting high quality and realism, easily fooling the human eye~\cite{karras2021alias, Rombach2022ldm, ramesh2022hierarchical, dalle3, firefly, imagine}.
Moreover, the newest image generation techniques allow users to alter images locally by combining parts of real and fake content, producing incredibly realistic manipulations~\cite{ramesh2022hierarchical, cleanupclipdrop}. 
While opening the doors to exciting scenarios, the widespread use of \gls{ai}-based image manipulation techniques poses significant challenges to the authenticity of digital content and public trust~\cite{cardenuto2023age}. Indeed, \gls{ai}-based image manipulation techniques can be easily exploited for malicious purposes such as spreading misinformation, performing identity theft and promoting fraudulent items and goods~\cite{deepfakesidentities, deepfakesonlinetrust}.

No less important, it has been proved that image tampering extends beyond digital multimedia content and it is increasingly affecting scientific images, i.e., images included in scientific publications~\cite{mirsky2019ct, qi2020emerging, gu2022ai}.
This issue poses a significant threat to the integrity and reliability of scientific research. An analysis of the scientific literature revealed that approximately $6\%$ of published papers contain manipulated images~\cite{bucci2018automatic}. Another study, which examined a large dataset of more than $20,000$ scientific papers, found that around $4\%$ of them included manipulated figures~\cite{bik2016prevalence}.

The latter study demonstrated that the manipulation issue is quite prevalent in scientific images depicting Western blots~\cite{bik2016prevalence}. 
Western blots, also known as western blotting or immunoblotting, are experimental techniques used to detect and quantify target proteins~\cite{blots}. 
They are extensively used in biomedical literature, particularly in molecular biology and immunogenetics, given their high sensitivity and precision. 

Standard manipulations of Western blots include the selective removal or addition of bands, targeted adjustments to brightness or contrast, and the elimination of background elements\cite{rossner2004s}. This issue is significant considering that Western blots have been used in approximately $8$ to $9\% $ of protein-related publications over the past $30$ years\cite{moritz202040}. 

Visual inspection is still the most common method for detecting manipulations in Western blot images. As a matter of fact, forensic techniques aimed at identifying local image tampering often struggle to detect manipulations in scientific images. This difficulty is often due to their lower pixel resolution and the numerous processing operations used to create realistic forgeries~\cite{sabir2021, Mandelli2022}.
Moreover, if the analyzed image has been synthetically manipulated, it would be almost impossible to detect the tampering traces through visual inspection~\cite{qi2020emerging}.
Within some preliminary experiments, the authors of~\cite{qi2020emerging} confirmed that conventional image generation methods that use \glspl{gan} could produce Western blots that were nearly indistinguishable from real ones, even to expert observers~\cite{isola2017image, park2019semantic}.
These findings highlight the critical need for increased vigilance to ensure the integrity of such images.

In this work, we tackle the task of image tampering localization in the context of Western blot images. 
We build upon the work of~\cite{Mandelli2022}, where the authors addressed the detection of synthetically generated Western blots through \glspl{gan} and \glspl{ddpm}. 
Here, we propose a method that is capable of localizing the region of insertion of synthetic content in pristine Western blot images. 

Our proposed approach works by analyzing small squared patches extracted from the query image. 
For a given image, we sequentially extract image patches and feed each patch to a detector trained to distinguish between real and synthetic content. 
By aggregating the scores from all patches, we produce a heatmap estimating the 
tampered region.

To assess our proposed methodology, 
we create and publicly release two datasets based on automatic and realistic manipulations applied to Western blots. In particular, realistic manipulations have been realized through classical image editing tools~\cite{gimp} as well as advanced \gls{ai}-based image manipulation models~\cite{ramesh2022hierarchical, cleanupclipdrop} that are not part of the training generators. 

Our proposed detector exhibits excellent performance results, being robust to unknown synthetic content generators and showing an almost total absence of false alarms.
We also compare with a recent state-of-the-art methodology developed for spotting image local manipulations~\cite{guillaro2023trufor}, discussing its advantages and limitations with respect to our proposal.
Eventually, we investigate our detector robustness in a challenging scenario, by testing it on manipulated
scientific images that belong to a completely different domain compared to Western blots, showing promising generalization capabilities.

\section{Localization of Synthetic Manipulations}
\label{sec:method}
In this section, we formulate the tackled task and we detail the proposed pipeline to address it.

\subsection{Problem Formulation}
\label{subsec:problem}

This paper proposes a method to localize the presence of synthetically generated content within pristine scientific images. 
Let us consider a manipulated image $\I$ with size $R \times C$, that presents a synthetically generated tampered region $\mathcal{T}$. We can associate $\I$ with a tampering mask $\mathbf{M}$ of size $R \times C$, where $\mathbf{M}$ is a 2D binary matrix in which each pixel takes a value of 0 if the corresponding pixel in $\I$ is pristine and a value of 1 if the pixel is synthetically generated. Formally, we define the pixel of $\mathbf{M}$ with coordinates $(r, c)$ as
\begin{equation}
    \label{eq:tampering-mask}
    \mathbf{M}(r,c) =
    \begin{cases}
      1 & \text{if}\ (r,c) \in \mathcal{T}, \\
      0 & \text{otherwise}.
    \end{cases}  
\end{equation}

\begin{figure}[t]
    \centering
    \includegraphics[width=\columnwidth]{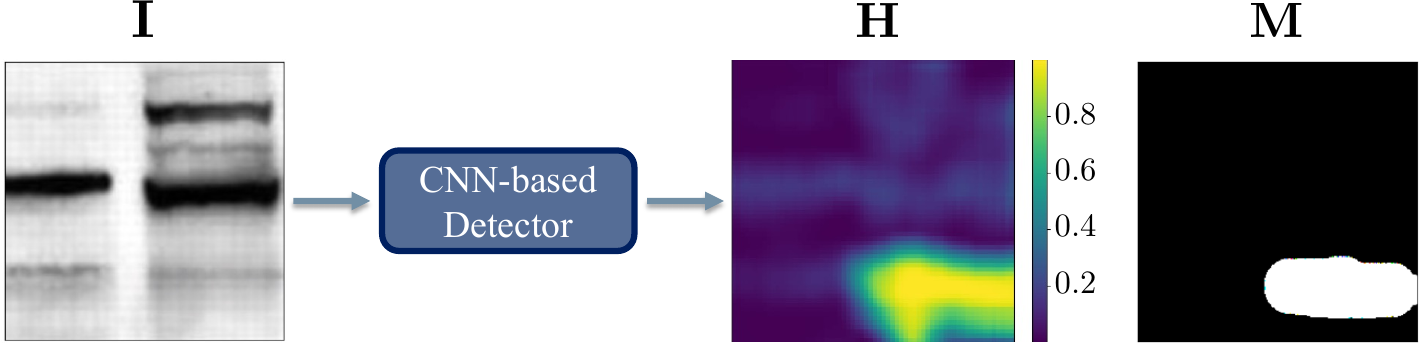}
    \caption[Problem Formulation]{Representation of the localization problem to identify regions of manipulation. The tampering mask $\mathbf{M}$ is shown as a reference. }
    \label{fig:problem-formulation}
\end{figure}

Our aim is to estimate a real-valued
tampering heatmap $\mathbf{H}$ as shown in Fig.~\ref{fig:problem-formulation}, 
which differentiates between real and synthetic regions, providing for each pixel the probability of it being synthetically generated. Thresholding $\mathbf{H}$ gives an estimate of $\mathbf{M}$.


\subsection{Proposed Methodology}
\label{subsec:method}

\begin{figure}[t]
    \centering
    \includegraphics[width=\columnwidth]{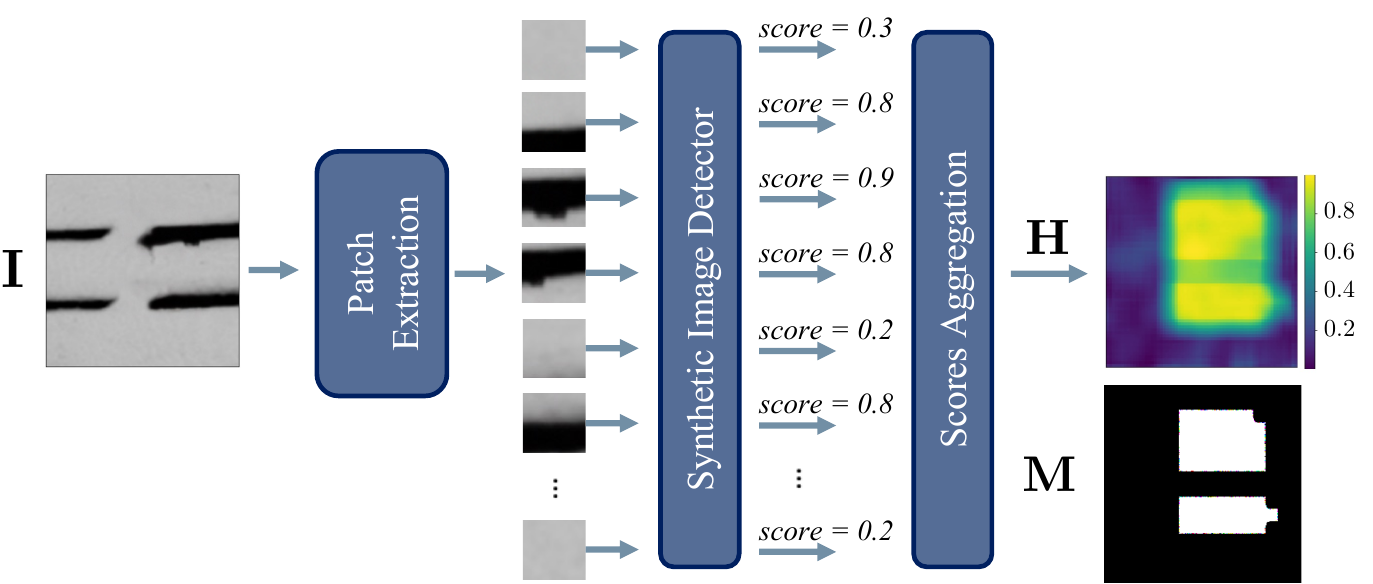}
    \caption[Pipeline of Proposed Methodology]{Pipeline of proposed methodology. We start from an image $\I$, whose patches are extracted sequentially and given as input to the synthetic image detector. The detector outputs probability scores that are aggregated to obtain the estimated heatmap $\mathbf{H}$, localizing the region of tampering. The tampering mask $\mathbf{M}$ is shown as a reference. }
    \label{fig:proposed-methodology}
\end{figure}

Our goal is to localize synthetically generated regions within scientific images. 
To do so, we propose to work in a patch-wise fashion, extracting and analyzing multiple patches from a single image under analysis. A sketch of the proposed methodology is shown in Fig.~\ref{fig:proposed-methodology}. Our pipeline is composed of three main steps:
\begin{enumerate}
    \item \textit{Patch extraction}, in which we sequentially extract small squared patches 
    from the image 
    under analysis. 
    \item \textit{\gls{cnn}-based detection}, in which every patch is analyzed by a \gls{cnn}-based detector trained to discriminate between genuine and synthetic content. 
    \item \textit{Heatmap estimation}, in which the patch scores are assembled to estimate a real-valued heatmap to tell real and synthetic pixels apart.
\end{enumerate}

\textbf{Patch extraction.} 
Given an image $\I$ under analysis, we sequentially extract small squared patches of $P \times P$ pixels from it. Every patch is defined as $\Patch_{\mathbf{I}_{i, j}}$, $(i, j)$ being the coordinates of its top-left corner pixel from the whole image $\I$. We consider an overlap between one patch and the next one, extracting patches with a specific stride of $S \times S$ pixels in both dimensions.
Notice that the analyzed image $\I$ can be of any size greater than $P \times P$ (which is very small), thus our proposed methodology can be potentially applied to both tiny and large-sized images.

\textbf{CNN-based detection.} 
Every extracted patch is analyzed by a \gls{cnn}-based synthetic image detector which differentiates between real and synthetic content. To do so, we select a straightforward detector that has been successfully exploited in the state-of-the-art for deepfake detection tasks~\cite{mandelli2020training, bonettini2021icpr, mandelli2022detecting}. 
In particular, we exploited the EfficientNet-B0 architecture which allows for a fast and lightweight training process. 
The detection network is trained on image patches $\Patch_{\mathbf{I}_{i, j}}$ and outputs a real score $ p_{i, j} \in [0, 1]$, where 0 means pristine and 1 synthetic (i.e., $p_{i, j}$ represents the likelihood of a patch being synthetically generated). 

To enhance robustness against post-processing operations, we employ strong data augmentations as recommended in several state-of-the-art studies on synthetic image detection~\cite{mandelli2022detecting}. 
These augmentations include horizontal and vertical flips, random 90-degree rotations, histogram equalization, random blurring, and random adjustments in brightness, contrast, color, and saturation. We also employ random downscaling and upscaling, and JPEG compression with quality factors randomly selected between 40 and 100. Each augmentation has a 50\% probability of being applied, except for JPEG compression, which has an 80\% probability. The parameters utilized are based on those outlined in~\cite{buslaev2020albumentations}.


\textbf{Heatmap estimation. }
We rearrange the detection scores according to their associated patch positions in the analyzed image, obtaining an estimated tampering heatmap $\mathbf{H}$. 
We associate each score $p_{i,j}$ with a patch of the heatmap $\Patch_{\mathbf{H}_{i, j}}$, placed in the same location of the analyzed patch $\Patch_{\mathbf{I}_{i, j}}$ in the query image. The patch $\Patch_{\mathbf{H}_{i, j}}$ holds constant pixel values all equal to $p_{i,j}$.
During the reconstruction procedure, heatmap patches are superimposed on each other based on the stride value. The probability scores are averaged in the regions where the patches overlap, resulting in a detailed prediction estimate. 

We believe the proposed methodology offers several advantages for detecting synthetic manipulations. Extracting small image patches allows the detector to be less influenced by potential biases related to the semantic content of the image. By examining very small pixel areas, the detector targets actual artifacts from the synthetic generation process rather than scene-specific characteristics. This feature is crucial for designing effective detectors that are robust to variations in the semantic content of test images. In our experiments, we demonstrate that our methodology exhibits notable generalization capabilities when tested on scientific images that have completely different characteristics from the training data.

\section{Experimental Setup}
\label{sec:setup}
Here, we present the datasets employed within this study and describe the training setup we used for the proposed real versus synthetic detector.

\subsection{Datasets}
\label{subsec:datasets}
\begin{figure}[t]
    \centering
        \includegraphics[width=.24\columnwidth]{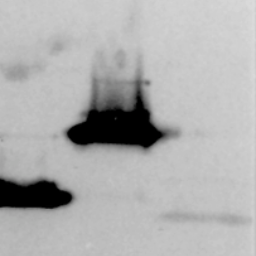}\hfill
        \includegraphics[width=.24\columnwidth]{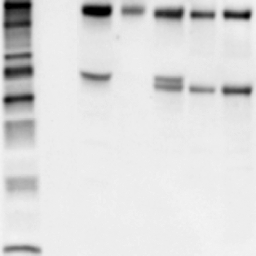}\hfill
        \includegraphics[width=.24\columnwidth]{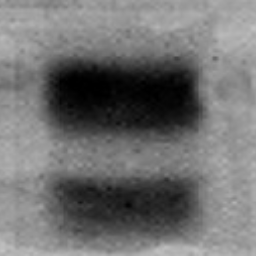}\hfill
        \includegraphics[width=.24\columnwidth]{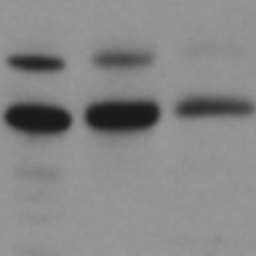}
    \caption{Pristine examples from the Western blots dataset released in~\cite{Mandelli2022}.}
    \label{fig:real-Western-blots-dataset}
\end{figure}

\begin{figure}
\centering
\setlength\tabcolsep{1pt}
    \begin{tabular}{c c c c}
        \text{\footnotesize{CycleGAN}} & \text{\footnotesize{\gls{ddpm}}} & \text{\footnotesize{Pix2pix}} & \text{\footnotesize{StyleGAN2-ADA}} \\
        \includegraphics[width=.24\columnwidth]{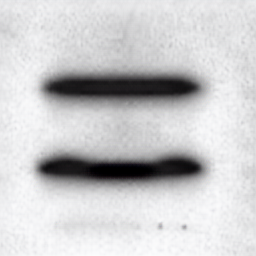} & \includegraphics[width=.24\columnwidth]{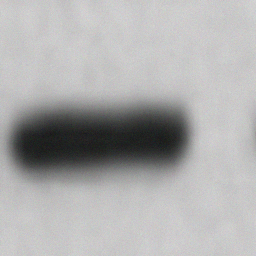} & \includegraphics[width=.24\columnwidth]{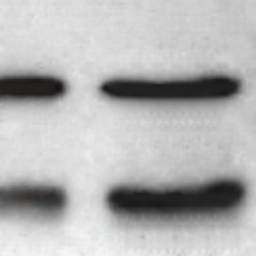} & \includegraphics[width=.24\columnwidth]{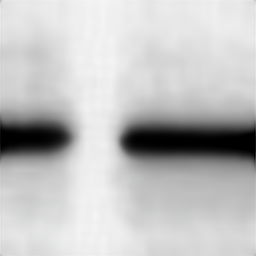}
    \end{tabular}
\caption{Synthetic examples from the Western blots dataset released in~\cite{Mandelli2022}.}
\label{fig:synth-Western-blots-dataset}
\end{figure}

\textbf{Western blots dataset. } 
We select scientific images from the Western blots dataset released in~\cite{Mandelli2022}. This dataset contains 14K real Western blot images and 24K synthetic images generated by four generative models (6K images per generator), namely CycleGAN \cite{zhu2017unpaired}, Pix2pix \cite{isola2017image}, StyleGAN2-ADA \cite{karras2020training}, and one \gls{ddpm} \cite{ho2020denoising}. 
All images have size $256 \times 256$ pixels. Some examples of the real and synthetic images are shown in Fig.~\ref{fig:real-Western-blots-dataset} and Fig.~\ref{fig:synth-Western-blots-dataset} respectively. 
We use $28$K images from this dataset equally split between real and synthetic image types to train and evaluate our detector. 

\textbf{Automatically manipulated Western blots. }
The dataset comprises $5690$ tampered Western blot images and has been built on top of the evaluation set (see Section~\ref{subsec:training_setup} for more details). 
We created the spliced images by inserting a synthetic squared patch of $64 \times 64$ pixels into a real host image at a random location. 
We extracted the patches from randomly chosen synthetically generated images belonging to the evaluation set. 
The amount of synthetic patches used is equally balanced between CycleGAN, Pix2pix, StyleGAN2-ADA and \gls{ddpm}.



\textbf{Realistically manipulated Western blots. } This dataset consists of $456$ manipulated Western blot images. As the automatically manipulated dataset, it is built upon the evaluation set of the Western blots dataset. 
We manipulated by hand pristine images in this dataset to emulate real-world forgery scenarios. To do so, we employed three different tampering tools, i.e., DALL$\cdot$E\,2~\cite{ramesh2022hierarchical}, Cleanup~\cite{cleanupclipdrop} and GIMP~\cite{gimp}. 

In all scenarios, we selected the area of tampering to perform insertion or deletion of Western blots. 
On GIMP this process was completely user-driven: we selected the pristine area and substituted it with synthetic content from the synthetic evaluation set to create realistic forgeries. 
DALL$\cdot$E\,2 and Cleanup tools are based on advanced diffusion models and enable the user to select a pixel region and substitute it with synthetically generated content. 

A single image can contain one or multiple local manipulations. On average, the images in this dataset are tampered by $4.12\%$, $13.93\%$, and $16.71\%$ for GIMP, Cleanup, and DALL$\cdot$E\,2 respectively. The percentages correspond to the number of tampered with pixels over the total pixels in the given image.

\textbf{M3Dsynth} \cite{zingarini2024m3dsynth}. This dataset consists of medical 3D \gls{ct} lung images with local manipulations. It consists of $8577$ images tampered using \glspl{gan} and Diffusion models working in 3D. 
Manipulations in M3Dsynth are performed by injecting or removing lung cancer nodules in real \gls{ct} scans. Images in this dataset are tampered with by extracting a 3D cube of size $32$mm from a real image, modifying the inner sides of the cube synthetically, and re-inserting it back in the real image. The authors exploited three different generation models to create the synthetic portions, namely a CycleGAN designed to work with 3D data \cite{iommi3dcyclegan}, a modified \gls{ddpm} \cite{dorjsembe2022three}, and CT-GAN~\cite{mirsky2019ct}. 

For our experiments, we consider the evaluation set of this dataset. This partition includes $205$ synthetically generated 3D images, each composed of multiple 2D slices with dimensions of $512 \times 512$ pixels.


\subsection{Synthetic Image Detector Configuration}
\label{subsec:training_setup}
The EfficientNet-B0 we employ within this study is pre-trained on the ImageNet dataset \cite{deng2009imagenet}.
We fine-tune it for the task of synthetic image detection on the Western blots dataset, splitting the dataset into training ($64\%$), validation ($16\%$) and test ($20\%$) partitions.
We use batches of size $250$, balanced in terms of real and synthetic samples, cross entropy loss, Adam optimizer, and a learning rate of $1 \times 10^{-3}$. We employ a scheduler to reduce the learning rate when the validation loss plateaus, with a minimum learning rate of $1 \times 10^{-8}$. A patience value of $50$ is used as a threshold for early stopping based on no improvements of the validation loss.


\section{Results}
\label{sec:results}
In this section, we present the results achieved by our proposed detection methodology.
We inspect the patch extraction parameters that best suit our synthetic detection task. We report our results on the automatically manipulated Western blots and on the realistically tampered ones. We perform an analysis of the false alarms returned by our detector on pristine Western blot images. We conclude this section by investigating the robustness of our methodology against the M3Dsynth dataset, which represents a challenging test set including different semantic content and different synthetic generation traces. 

\subsection{Parameters Selection for Synthetic vs Real Detection}
\label{subsec:results_parameter_synthetic}

In the patch extraction phase, we select patches of size $P \times P$ pixels, investigating different values $P \in [32, 64, 96, 128]$. 
This results in four synthetic versus real detectors, each having a different input patch size. 
To define the best patch size for our task, we test the four versions of our synthetic detector on the test partition of the Western blots dataset (extracting all the non-overlapping patches of $P \times P$ pixels from the testing images). 

Table~\ref{tab:patch_size_analysis} reports the maximum \gls{ba} in classifying real versus synthetic content achieved for all the considered patch sizes. 
As expected, larger patch sizes yield better results since the model receives more information for predictions. Nevertheless, all configurations exhibit robust performance, with only a slight decrease as the patch size is reduced (we pass from a maximum \gls{ba} of $99.8\%$ to $96.5\%$). Based on these findings, we perform the next experiments using a patch size of $32 \times 32$ pixels. We believe a smaller patch size is more suitable for the task at hand, as it allows us to detect small-sized manipulations. 

\begin{table}
\centering 
\caption{Maximum BA achieved for synthetic versus real detection on the test set of the Western blots dataset across the considered patch sizes. In bold, the best results. }
\label{tab:patch_size_analysis}
\resizebox{.52\columnwidth}{!}{
\begin{tabular}{cc}
\toprule 
\textbf{Patch size} $P \times P$  & \textbf{Max BA} \\ \midrule
$128 \times 128$    & $\mathbf{99.8\%}$   \\ 
$96 \times 96$      & $99.3\%$   \\
$64 \times 64$      & $99.2\%$   \\
$32 \times 32$      & $96.5\%$   \\   \bottomrule
\end{tabular}
}
\end{table}

\subsection{Performance on Automatically Manipulated Western blots}
\label{subsec:results_automatic}
\begin{figure}[t]
\centering
\setlength\tabcolsep{1pt}
    \begin{tabular}{c c c c}
        \text{\footnotesize{Pristine image}} & \text{\footnotesize{Manipulated image}} & \text{\footnotesize{Tampering mask}} & \text{\footnotesize{Estimated heatmap}} \\
        \includegraphics[width=.24\columnwidth]{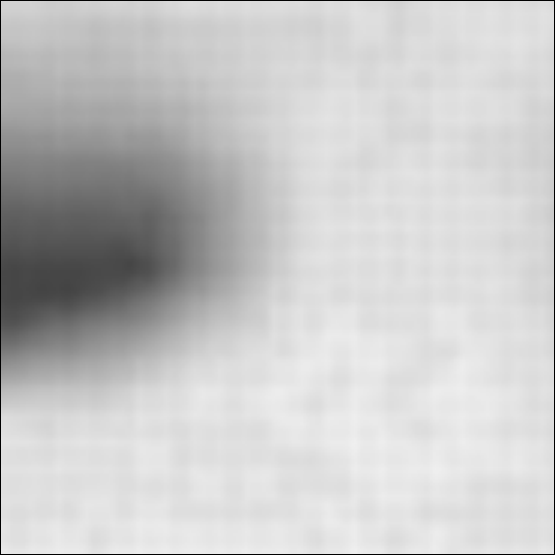} & \includegraphics[width=.24\columnwidth]{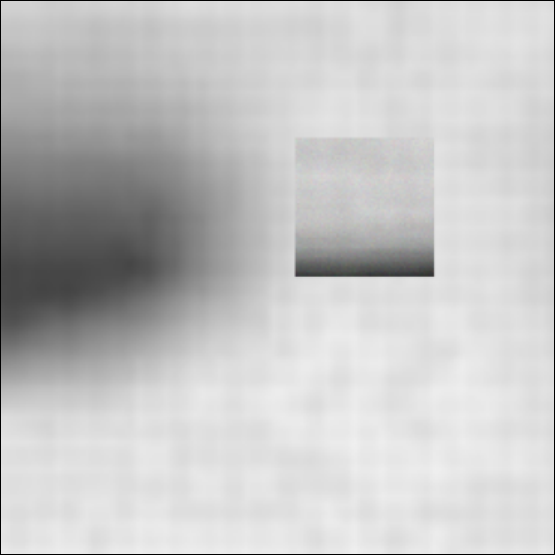} & \includegraphics[width=.24\columnwidth]{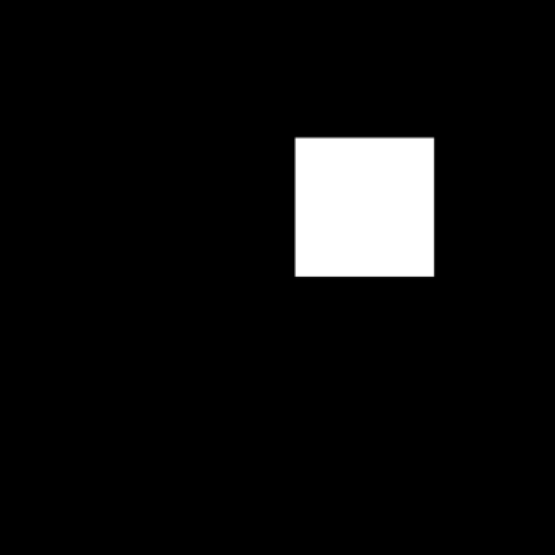} & \includegraphics[width=.24\columnwidth]{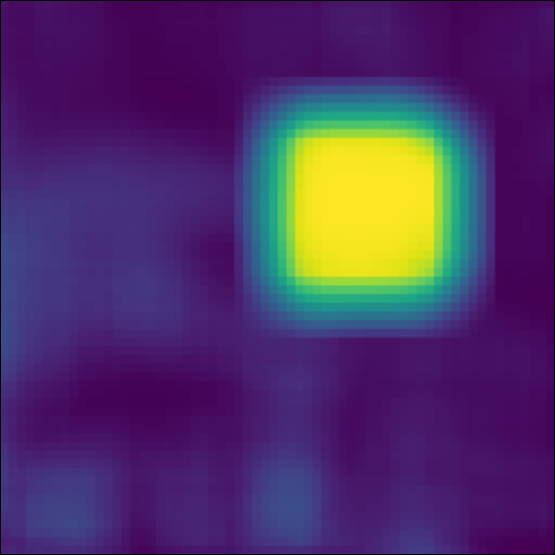} \\
        \includegraphics[width=.24\columnwidth]{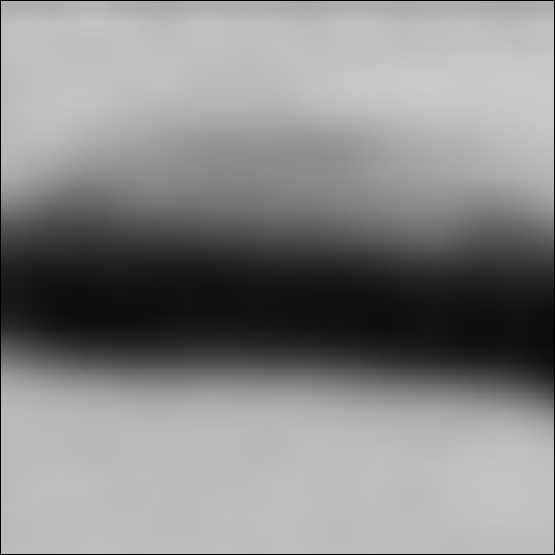} & \includegraphics[width=.24\columnwidth]{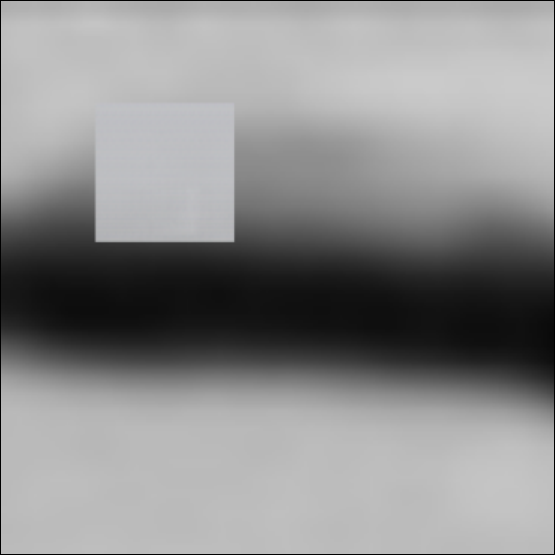} & \includegraphics[width=.24\columnwidth]{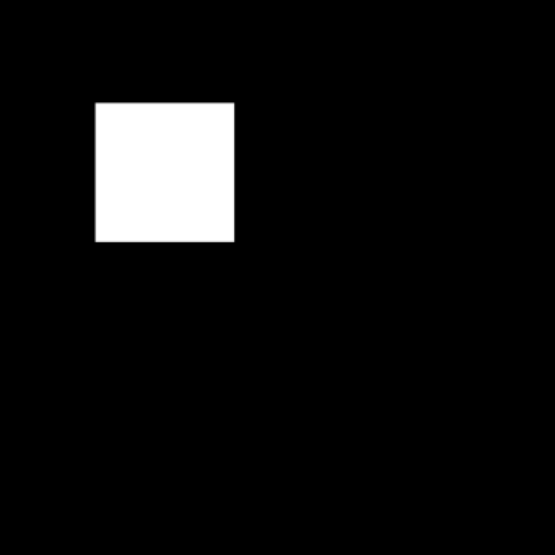} & \includegraphics[width=.24\columnwidth]{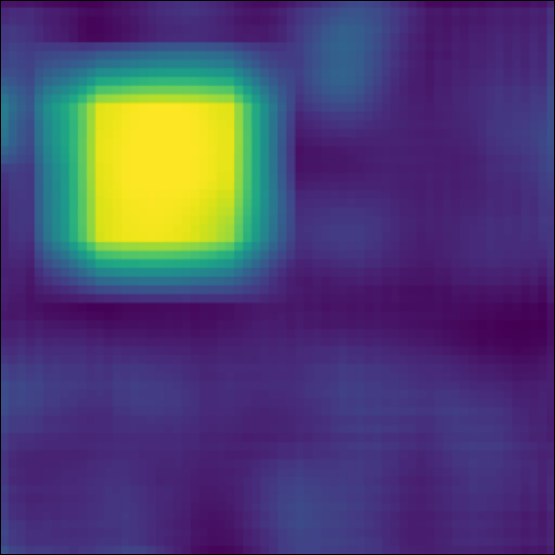} \\
        \includegraphics[width=.24\columnwidth]{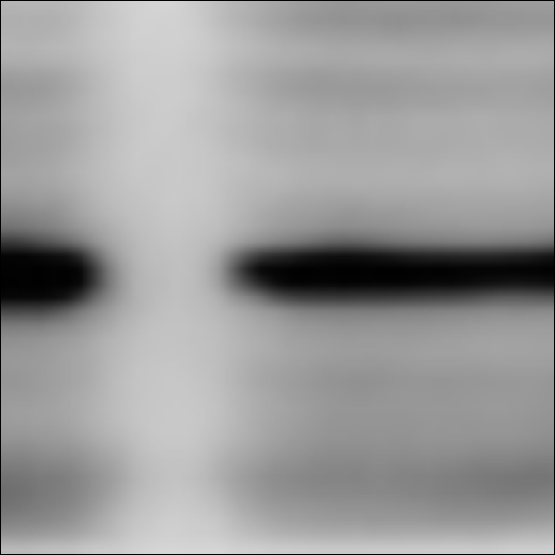} & \includegraphics[width=.24\columnwidth]{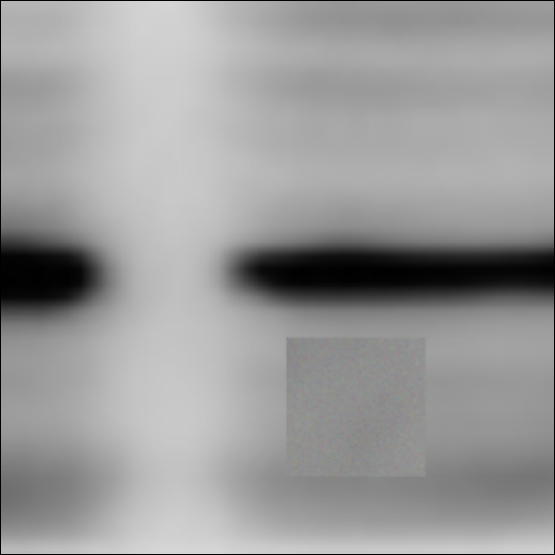} & \includegraphics[width=.24\columnwidth]{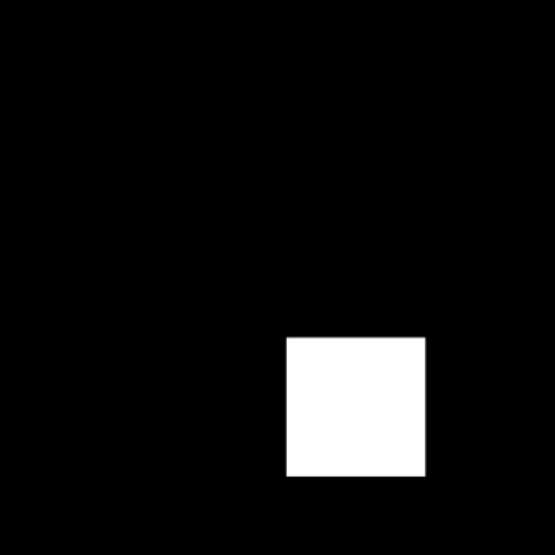} & \includegraphics[width=.24\columnwidth]{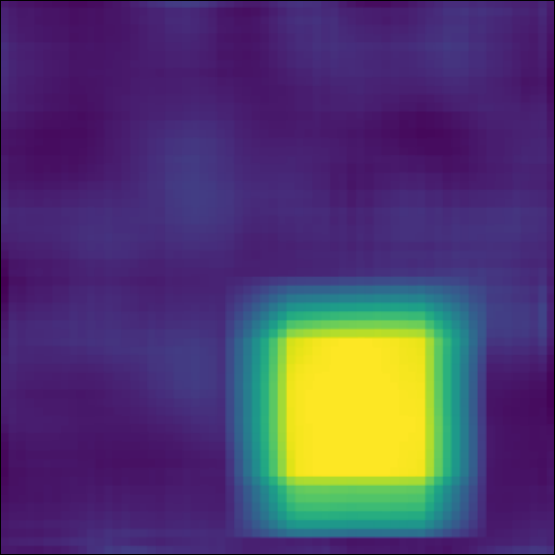}
    \end{tabular}
\caption{Localization results on automatically tampered Western blot images, for a patch size of $32 \times 32$ pixels and stride $4 \times 4$. }
\label{fig:automatic_manipulations}
\vspace{-.5em}
\end{figure}

In this section, we assess the performance of the model in detecting and localizing the tampering region in automatically manipulated Western blot images. A few examples of these images are shown in the second column of Fig.~\ref{fig:automatic_manipulations}. These are not meant to be realistic images (as the other datasets we are using), but help providing sufficient statistical results.
As specified before, we always extract patches of size $32 \times 32$ pixels from the image under analysis. We experiment with various stride values of $S \times S$ pixels, considering $S \in [4, 8, 16, 32]$.

\begin{table}[t]
\centering 
\caption{AUC and maximum BA values achieved on the automatically tampered Western blot images for different stride values. In bold, the best results per metrics.}
\label{tab:aut_tamp_results}
\resizebox{.6\columnwidth}{!}{
\begin{tabular}{ccc}
\toprule 
\textbf{Stride} $S \times S$     &\textbf{AUC}   & \textbf{Max BA} \\ \midrule
$32 \times 32$                  & $0.922$         & $91.8\% $  \\ 
$16 \times 16$                  & $0.974$         & $95.1\%$   \\
$8 \times 8$                   & $0.981$         & $95.9\%$   \\
$4 \times 4$                   & $\mathbf{0.984}$         & $\mathbf{96.2\%}$   \\   \bottomrule
\end{tabular}
}
\end{table}

Table~\ref{tab:aut_tamp_results} reports the results in terms of the \gls{auc} and the maximum \gls{ba} achieved at different stride values. 
In this case, both metrics refer to the comparison between a single estimated heatmap and its ground-truth tampering mask. We report the average metrics computed over the considered test set. 

Results show that lower stride values lead to more accurate heatmap estimates, which help in localizing tampered regions. 
This outcome was expected since lower strides allow to exploit multiple patches contributions for estimating the heatmap values. 
Based on these outcomes, a stride value of $4 \times 4$ pixels is chosen for further experiments. 

In particular, Fig.~\ref{fig:automatic_manipulations} depicts examples of pristine Western blot images from the test set, their related manipulated versions with the corresponding tampering masks and the estimated heatmaps. 
It is worth noticing the strong correspondence between the heatmap and the tampering mask, meaning for high detection rate in locating the tampering areas and extremely reduced false alarms (i.e., pristine pixels are associated with low heatmap values).  


\begin{figure}
    \centering
    \includegraphics[width=\columnwidth]{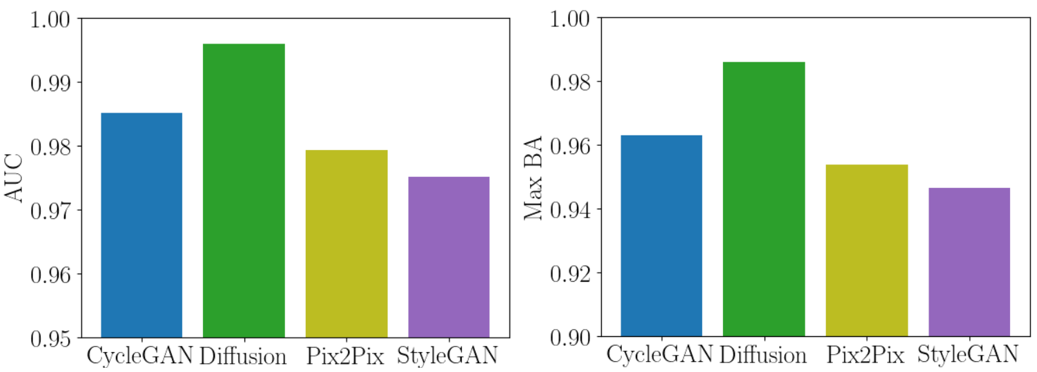}
    \caption[Generators Comparison for Synthetic Image Detection]{Performance in the automatically tampered images across each generator by AUC and maximum BA.}
    \label{fig:generators}
    \vspace{-1em}
\end{figure}

We further investigate the localization performance of our detector by analyzing the results on each generator used for the local manipulations. Results in Fig.~\ref{fig:generators} show that our detector performs remarkably well for all generators (\gls{auc} $> 0.97$ and maximum \gls{ba} $> 0.94$).
Moreover, the detector finds it easiest to identify synthetic tamperings from the \gls{ddpm} model (diffusion-based). This reinforces the findings in~\cite{Mandelli2022} where a similar result was observed.

\subsection{Performance on Realistically Tampered Western blots}
\label{subsec:results_realistic}

\begin{figure}[t]
    \centering
    \begin{subfigure}[b]{\columnwidth}
        \centering
        \setlength\tabcolsep{.3pt}
        \begin{tabular}{c c c c}
        {\footnotesize{Pristine image}} & {\footnotesize{Manipulated image}} & {\footnotesize{Tampering mask}} & {\footnotesize{Estimated heatmap}} \\
        \includegraphics[width=.24\columnwidth]{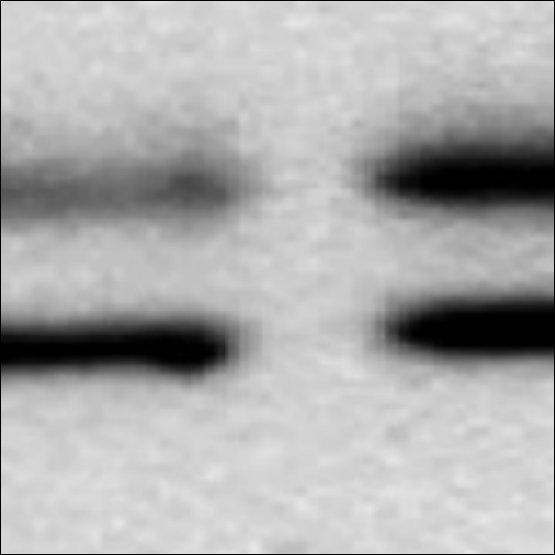} &
        \includegraphics[width=.24\columnwidth]{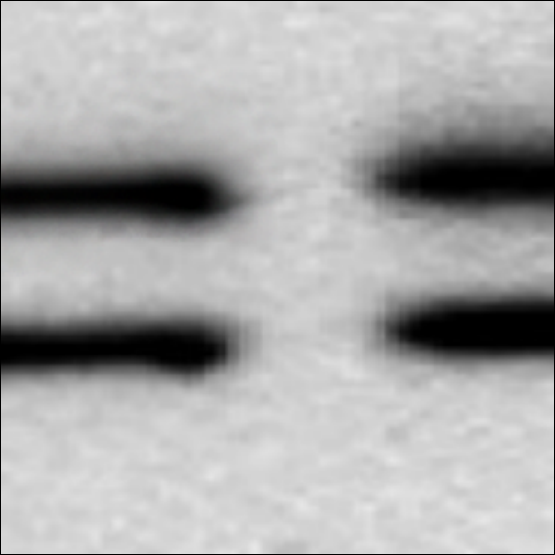} &
        \includegraphics[width=.24\columnwidth]{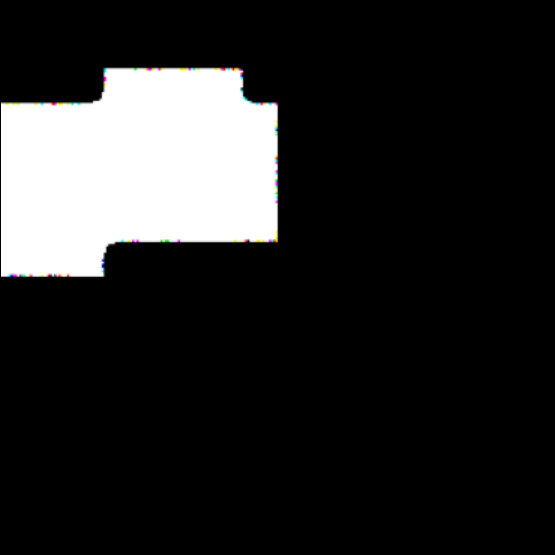} &
        \includegraphics[width=.24\columnwidth]{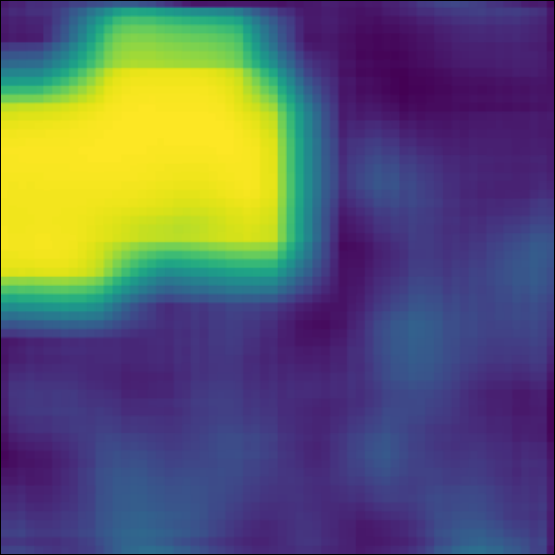} \\
        \includegraphics[width=.24\columnwidth]{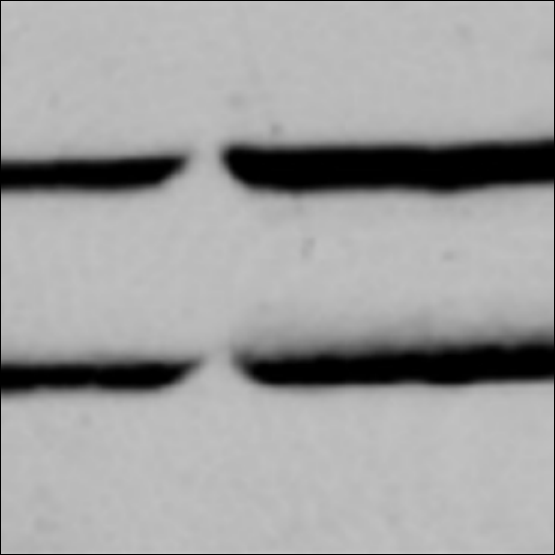} &
        \includegraphics[width=.24\columnwidth]{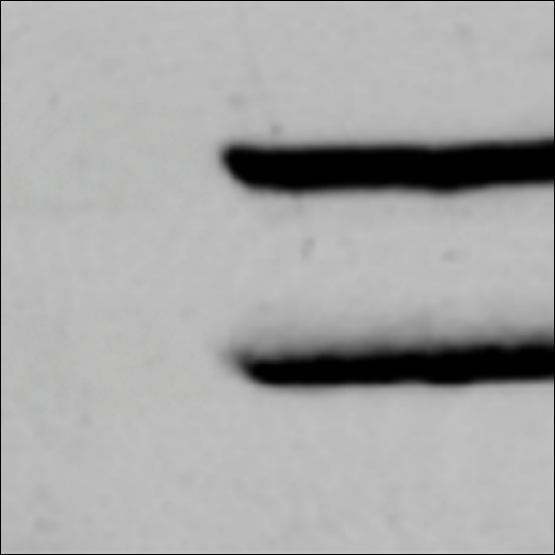} &
        \includegraphics[width=.24\columnwidth]{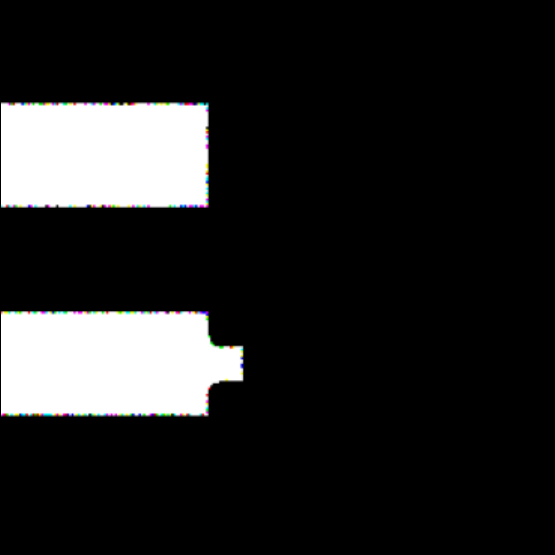} &
        \includegraphics[width=.24\columnwidth]{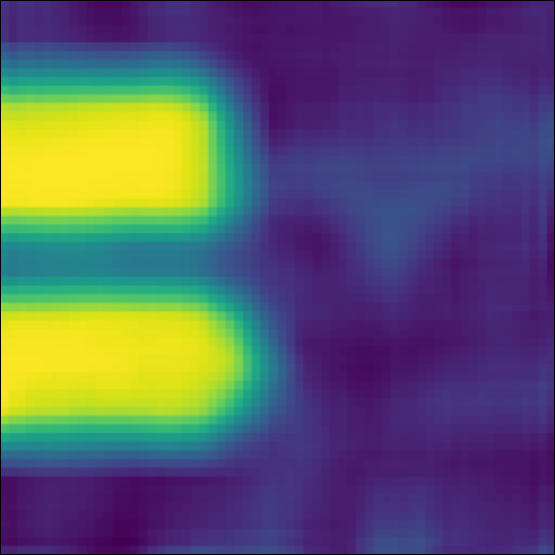} 
        \end{tabular}
        \caption{Realistic manipulations realized with DALL$\cdot$E\,2.}
        \label{subfig:dalle}
    \end{subfigure}
    \begin{subfigure}[b]{\columnwidth}
        \centering
        \setlength\tabcolsep{.3pt}
        \begin{tabular}{c c c c}
        {\footnotesize{Pristine image}} & {\footnotesize{Manipulated image}} & {\footnotesize{Tampering mask}} & {\footnotesize{Estimated heatmap}} \\
        \includegraphics[width=.24\columnwidth]{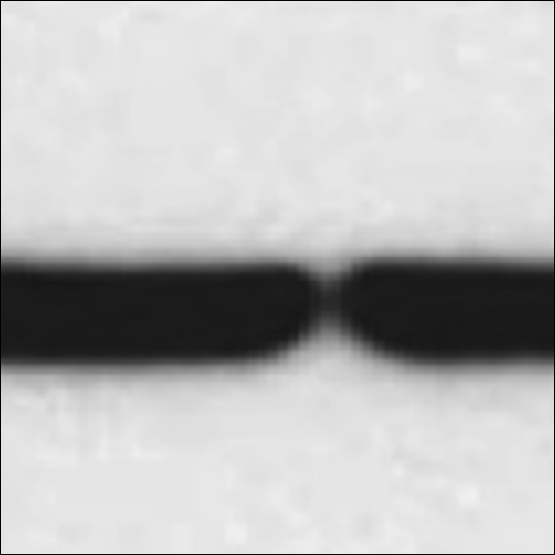} &
        \includegraphics[width=.24\columnwidth]{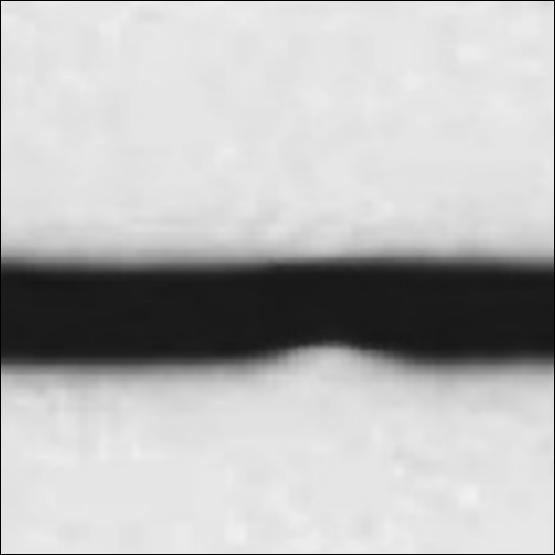} &
        \includegraphics[width=.24\columnwidth]{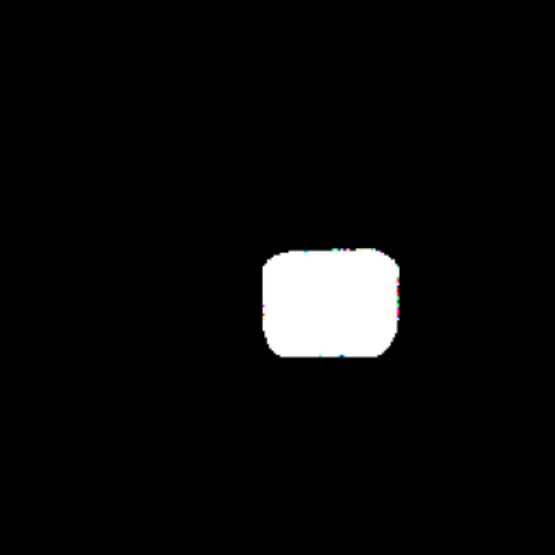} &
        \includegraphics[width=.24\columnwidth]{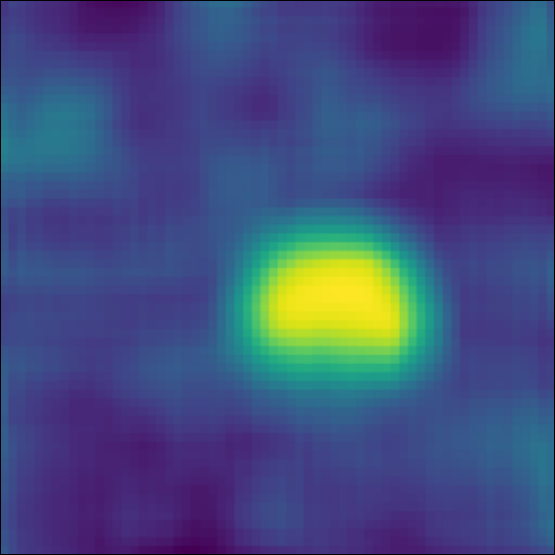} \\
        \includegraphics[width=.24\columnwidth]{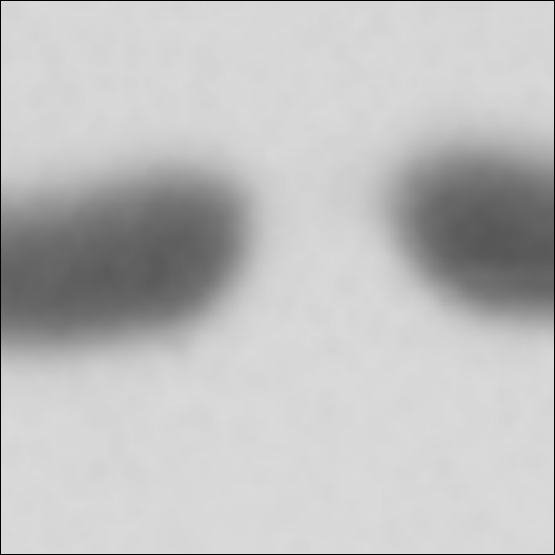} &
        \includegraphics[width=.24\columnwidth]{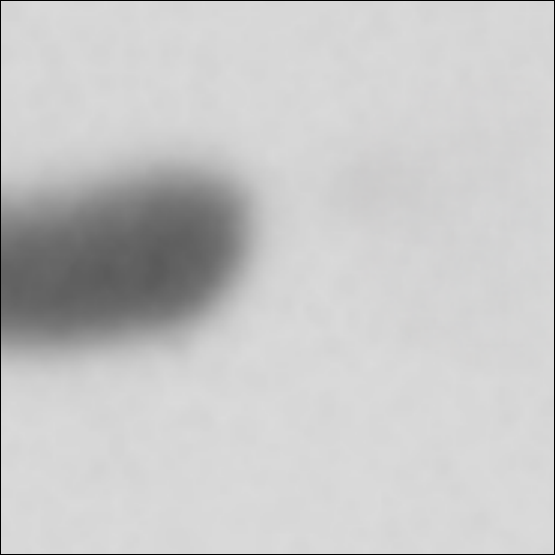} &
        \includegraphics[width=.24\columnwidth]{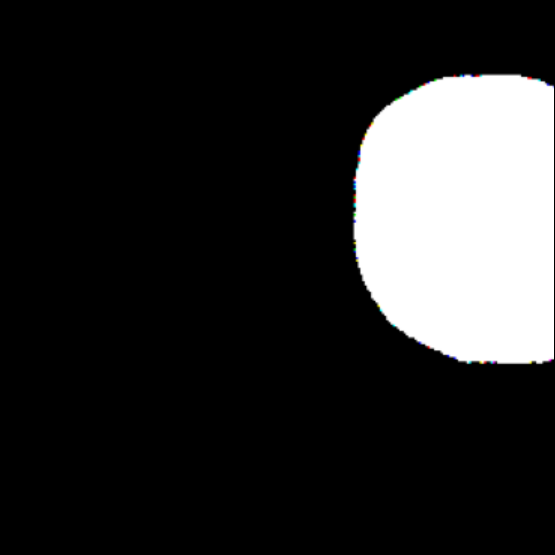} &
        \includegraphics[width=.24\columnwidth]{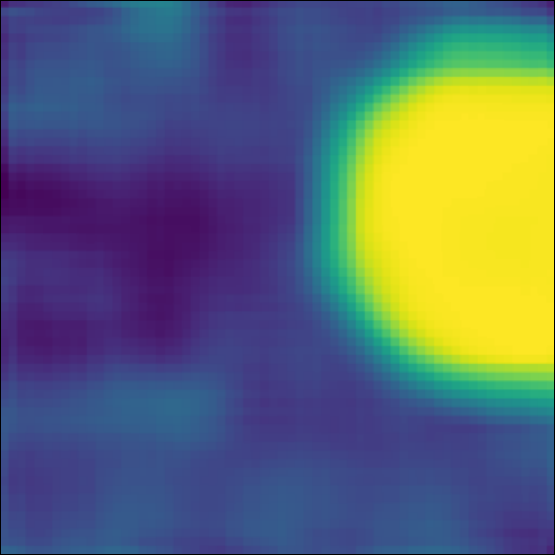}
        \end{tabular}
        \caption{Realistic manipulations realized with Cleanup.}
        \label{subfig:cleanup}
    \end{subfigure}
    \begin{subfigure}[b]{\columnwidth}
        \centering
        \setlength\tabcolsep{.3pt}
        \begin{tabular}{c c c c}
        {\footnotesize{Pristine image}} & {\footnotesize{Manipulated image}} & {\footnotesize{Tampering mask}} & {\footnotesize{Estimated heatmap}} \\
        \includegraphics[width=.24\columnwidth]{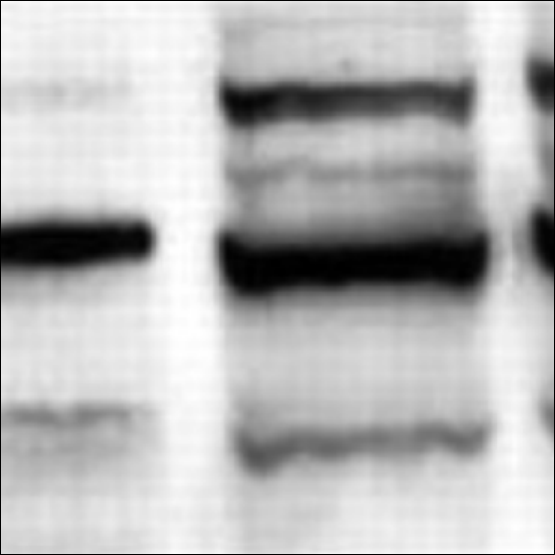} &
        \includegraphics[width=.24\columnwidth]{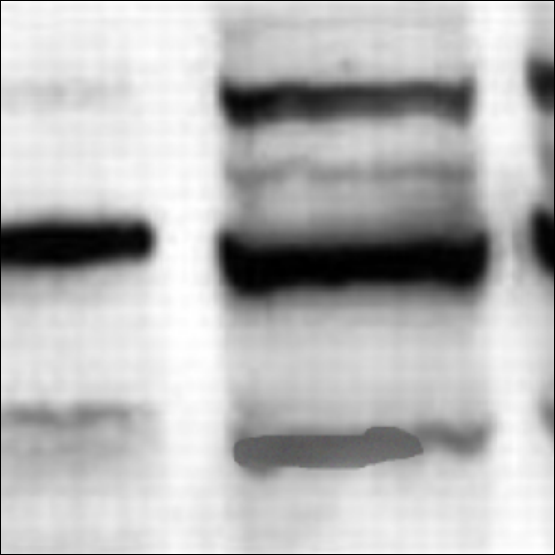} &
        \includegraphics[width=.24\columnwidth]{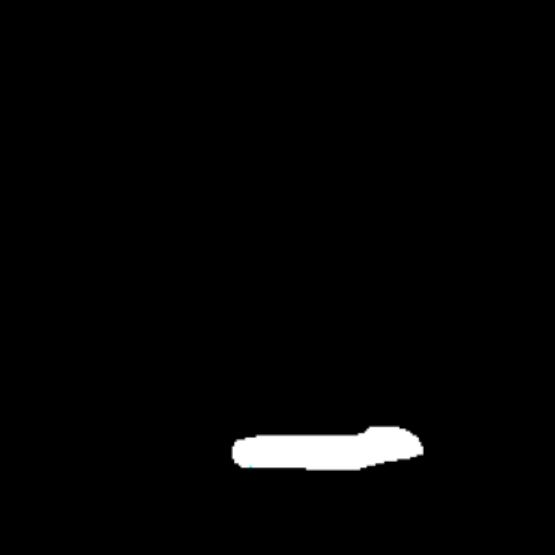} &
        \includegraphics[width=.24\columnwidth]{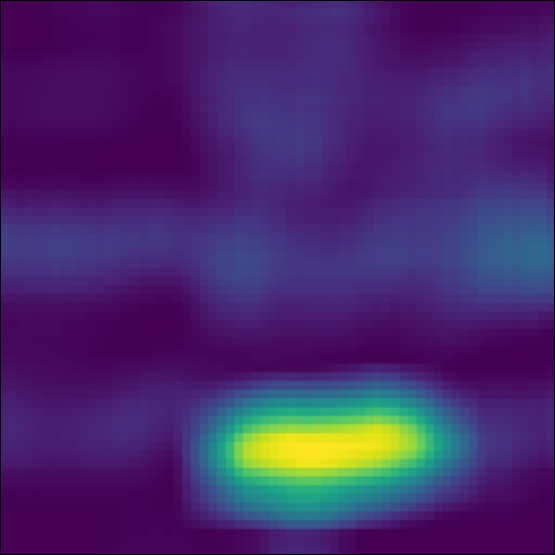} \\
        \includegraphics[width=.24\columnwidth]{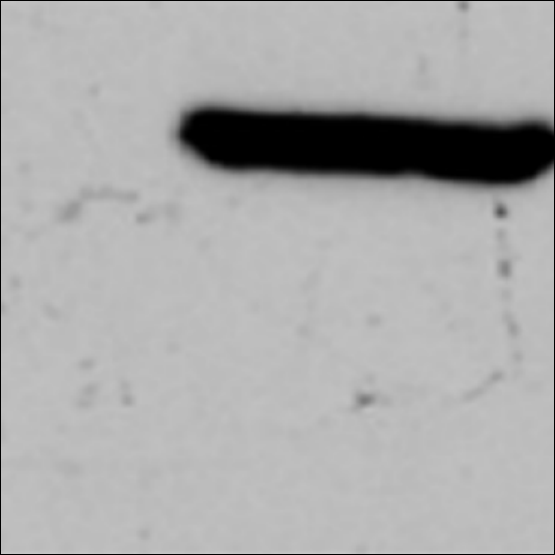} &
        \includegraphics[width=.24\columnwidth]{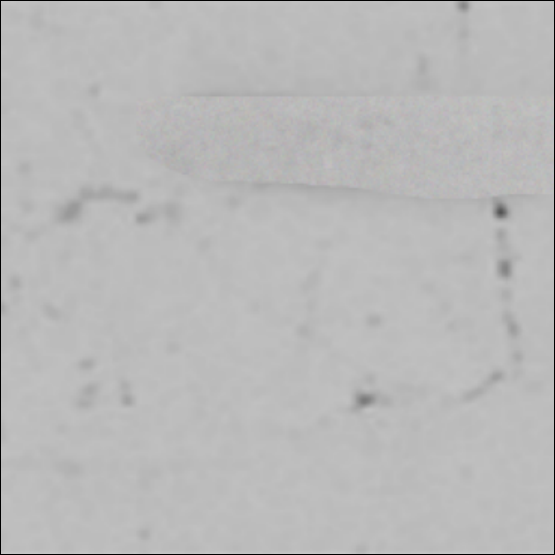} &
        \includegraphics[width=.24\columnwidth]{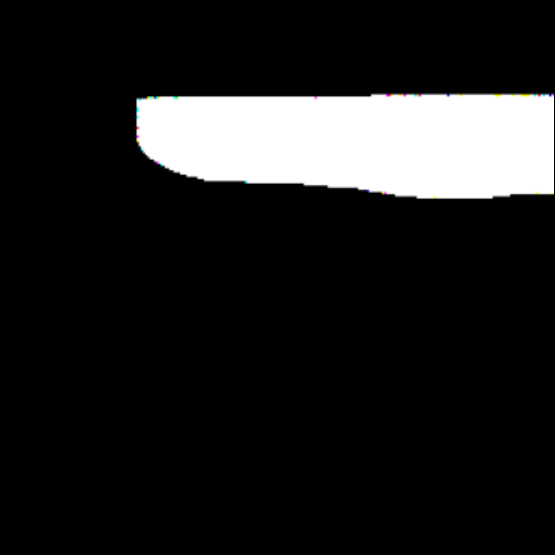} &
        \includegraphics[width=.24\columnwidth]{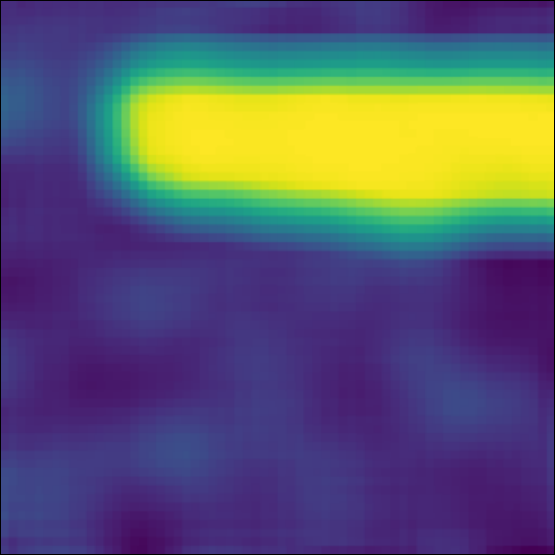}
        \end{tabular}
        \caption{Realistic manipulations realized with GIMP.}
        \label{subfig:gimp}
    \end{subfigure}
    \caption{Examples of localization results on realistically tampered images showing both insertion and deletion of Western blots across different tampering methods.}
    \label{fig:realistic}
\end{figure}

Here, we test our detector on realistically tampered Western blot images. Some examples are shown in Fig.~\ref{fig:realistic}, for all three different tampering methods. 

To validate our results, we benchmark against a state-of-the-art model recently developed for the localization of image manipulations. 
The selected baseline is the TruFor method~\cite{guillaro2023trufor}, which is based on a transformer network trained on a huge amount of images, undergone many different post-processing. It has demonstrated to achieve excellent performances for image forgery detection and localization by integrating RGB data with noise-sensitive fingerprints. 

Table~\ref{tab:real_tamp_results} shows \gls{auc} and maximum \gls{ba} results for each manipulation technique. 
Our detector performs well in localizing forgeries across all tampering methods, with an average \gls{auc} of $0.976$ and an average maximum \gls{ba} of $95.8\%$.
It is worth noticing that our detector was not trained over synthetic data generated through DALL$\cdot$E\,2 and Cleanup models, however we demonstrate excellent performances on them. This might be due to similar generation artifacts of these diffusion-based generators with those left by the \gls{ddpm} used in the training phase. 
The TruFor method proves the best solution in case of classical image manipulations like GIMP-made ones. Nonetheless, we achieve a comparable \gls{auc} and we lose only two percentage points in the maximum \gls{ba}.
While TruFor obtains acceptable localization results on DALL$\cdot$E\,2, it suffers more on Cleanup-based forgeries, which reveal challenging to detect.
These worse results of the TruFor solution might be related to the completely different training setup of the method, which is never trained on synthetic content but it is trained to detect subtle editing differences in local residual patterns of the analysed images. 

\begin{table}
\centering 
\large
\caption{AUC values and maximum BA achieved by our method and TruFor~\cite{guillaro2023trufor} on realistically manipulated Western blots. In bold, the best results per metrics and per manipulation method. }
\label{tab:real_tamp_results}
\resizebox{\columnwidth}{!}{\begin{tabular}{ccccccc}
\toprule
& \multicolumn{2}{c}{\textbf{DALL$\cdot$E\,2}} & \multicolumn{2}{c}{\textbf{Cleanup}} & \multicolumn{2}{c}{\textbf{GIMP}} \\ \midrule
\multicolumn{1}{l}{} & AUC              & Max BA           & AUC             & Max BA           & AUC            & Max BA         \\ \midrule 
\textbf{Proposed}    & $\mathbf{0.991}$   & $\mathbf{97.5\%}$    & $\mathbf{0.973}$  & $\mathbf{95.4\%}$    &$ 0.970  $        & $95.0\%$           \\
\textbf{TruFor}      & $0.845$            & $84.9\% $            & $0.490$            & $60.4\% $            & $\mathbf{0.975}$ & $\mathbf{97.3\%}$    \\ \bottomrule
\end{tabular}}
\end{table}



\begin{figure}[t]
    \centering
    \subfloat[Selected tampering region]{\includegraphics[width=0.3\columnwidth]{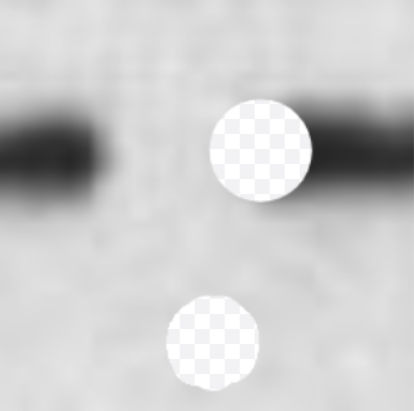}
    \label{fig:dalle2-square-inpainting-tampering-region}}
    \subfloat[Tampering mask]{\includegraphics[width=0.3\columnwidth]{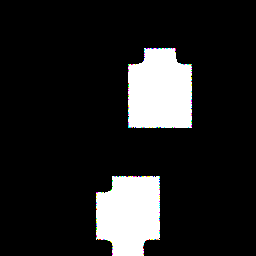}
    \label{fig:dalle2-square-inpainting-mask}}
    \caption{Example of difference between the tampering region selected by the user and the tampering mask operated by DALL$\cdot$E\,2.}
    \label{fig:dalle2-square-inpainting}
\end{figure}

As a side note, DALL$\cdot$E\,2 is noted to generate square-shaped manipulations regardless of the original shape provided by the user. An example of this behaviour observed for circular-shape manipulations is shown in Fig.~\ref{fig:dalle2-square-inpainting}. 

\subsection{Handling False Alarms}
\label{subsec:results_false_alarms}
We conduct an additional experiment to assess the behavior of our detector in absence of manipulations, i.e., we test it over pristine Western blots and we check if it is keen to return false alarms. 
To this purpose, we compute the threshold associated with the maximum \gls{ba} achieved over the realistically tampered Western blots dataset, which corresponds to heatmap values equal to $0.736$. 
Pixels with detection scores below the threshold are classified as real, while those with scores greater than or equal to the threshold are classified as synthetic. 

As a test set, we consider a subset of pristine Western blots from the evaluation partition that have not been used to generate realistically tampered images. This subset comprises $1950$ genuine images. Considering the threshold computed above, we compute the correct detection rate as the number of pixels correctly classified as real vs the total amount of pixels,
obtaining a final value of $99.3\%$. 
These findings suggest that we can guarantee a low rate of false alarms when our detector is presented with authentic images.

\begin{table}[t]
\centering 
\caption{AUC and maximum BA values achieved on M3Dsynth for different generation techniques.}
\label{tab:m3dsynth}
\resizebox{.6\columnwidth}{!}{
\begin{tabular}{ccc}
\toprule 
\textbf{Generation Technique}     &\textbf{AUC}   & \textbf{Max BA} \\ \midrule
3D CycleGAN       & $0.704$        & $67.5\%$   \\ 
3D \gls{ddpm}     & $0.749 $       & $70.9\%$   \\
CT-GAN       & $0.764$        & $71.3\%$   \\   \bottomrule
\end{tabular}
}
\end{table}

\subsection{Performance on M3Dsynth}
\label{subsec:results_4}
In this section, we test our proposed method in a challenging scenario, assessing our generalization capabilities on the M3Dsynth dataset.

In Table~\ref{tab:m3dsynth} we report the \gls{auc} and the maximum \gls{ba} with respect to the different generators used to create the synthetic patches.
While the detector preserves some robustness, results show an evident drop in its performance with respect to previous experiments. This was indeed an expected outcome for two main reasons. 
First, M3Dsynth images and Western blots completely differ in terms of semantics. Second, the synthetic generation techniques are different from those included in our training set: some of the generators have been seen only in their 2D versions (CycleGAN and \gls{ddpm}), others are completely unknown (CT-GAN). 

\section{Conclusions}
\label{sec:conclusions}

In this work, we address the task of localizing synthetic tampering regions within Western blot images.
To tackle this issue, we propose a method working in a patch-wise fashion: we iteratively extract small image patches from the image under analysis, and we pass each patch through a synthetic vs real detector providing the likelihood of every patch being synthetically generated.
Eventually, we aggregate patches' scores and build a tampering heatmap that reveals the image area in which the manipulation has been performed.

To assess our performances, we created two different evaluation datasets. One dataset encompasses automatically generated spliced images, while the other contains manually performed manipulations using advanced \gls{ai}-based image manipulation tools that were not known at training phase. Both dataset have been publicly released to allow for future comparisons and investigations by the forensic community.  

The proposed method achieves excellent performances on both automatically and realistically tampered with Western blots, exhibiting an extremely reduced false alarms rate. We also compare our results with a state-of-the-art methodology, discussing advantages and limitations of both solutions. Additional testing for robustness on the M3Dsynth dataset~\cite{zingarini2024m3dsynth}, which contains completely different semantics and manipulations, yields promising results.


\bibliographystyle{IEEEtran}
\bibliography{references}

\end{document}